\documentclass[lettersize,journal]{IEEEtran}
\usepackage{amsmath,amsfonts}
\usepackage{algorithmic}
\usepackage{array}
\usepackage[caption=false,font=normalsize,labelfont=sf,textfont=sf]{subfig}
\usepackage{textcomp}
\usepackage{stfloats}
\usepackage{url}
\usepackage{verbatim}
\usepackage{graphicx}
\usepackage{bm}
\usepackage{hyperref}
\usepackage{cleveref}

\newtheorem{theorem}{Theorem}

\newtheorem{corollary}{Corollary}

\usepackage{tikz}

\newcommand\submittedtext{%
  \footnotesize This work has been submitted to the IEEE for possible publication. Copyright may be transferred without notice, after which this version may no longer be accessible.}

\newcommand\submittednotice{%
\begin{tikzpicture}[remember picture,overlay]
\node[anchor=south,yshift=10pt] at (current page.south) {\fbox{\parbox{\dimexpr0.65\textwidth-\fboxsep-\fboxrule\relax}{\submittedtext}}};
\end{tikzpicture}%
}

\def\BibTeX{{\rm B\kern-.05em{\sc i\kern-.025em b}\kern-.08em
    T\kern-.1667em\lower.7ex\hbox{E}\kern-.125emX}}
\usepackage{balance}

\usepackage{xcolor}    
\usepackage{cite} 
\usepackage{import}
\usepackage{amsmath}

\begin{document}
\title{Safe Uncertainty-Aware Learning of Robotic Suturing}
\author{
Wilbert Peter Empleo,
Yitaek Kim,
Hansoul Kim,
Thiusius Rajeeth Savarimuthu,
Iñigo Iturrate

\thanks{This work was supported by the Active Uncertainty-aware Robot Imitation (AURI) project, supported by the Fabrikant Vilhelm Pedersen og Hustrus Legat.}
\thanks{Wilbert Peter Empleo, Yitaek Kim, Thiusius Rajeeth Savarimuthu, and Iñigo Iturrate are with SDU Robotics, The Maersk Mc-Kinney Moller Institute, University of Southern Denmark, Odense, Denmark.}
\thanks{Hansoul Kim is with the Department of Mechanical Engineering, Myongji University, Yongin-si, Gyeonggi-do, Republic of Korea.}
}

\maketitle

\begin{abstract}
Robot-Assisted Minimally Invasive Surgery is currently fully manually controlled by a trained surgeon. Automating this has great potential for alleviating issues, e.g., physical strain, highly repetitive tasks, and shortages of trained surgeons. For these reasons, recent works have utilized Artificial Intelligence methods, which show promising adaptability. Despite these advances, there is skepticism of these methods because they lack explainability and robust safety guarantees.
This paper presents a framework for a safe, uncertainty-aware learning method. We train an Ensemble Model of Diffusion Policies using expert demonstrations of needle insertion. Using an Ensemble model, we can quantify the policy's epistemic uncertainty, which is used to determine Out-Of-Distribution scenarios. This allows the system to release control back to the surgeon in the event of an unsafe scenario. Additionally, we implement a model-free Control Barrier Function to place formal safety guarantees on the predicted action.  
We experimentally evaluate our proposed framework using a state-of-the-art robotic suturing simulator. We evaluate multiple scenarios, such as dropping the needle, moving the camera, and moving the phantom. 
The learned policy is robust to these perturbations, showing corrective behaviors and generalization, and it is possible to detect Out-Of-Distribution scenarios. We further demonstrate that the Control Barrier Function successfully limits the action to remain within our specified safety set in the case of unsafe predictions.
\end{abstract}


\submittednotice

\section{Introduction}
\IEEEPARstart{R}{ecent} research using artificial intelligence (AI) for autonomous robot-assisted minimally-invasive surgery (RMIS) has shown great promise in terms of generalizability and adaptability \cite{kim2024surgicalrobottransformersrt,yu2024orbitsurgicalopensimulationframeworklearning,wu2025surgicaihierarchicalplatformfinegrained}. This is important, as currently RMIS is fully manually teleoperated by a surgeon, which due to the shortage of trained surgeons not only limits the accessibility of advanced surgical procedures but also places a considerable burden on the existing surgical workforce \cite{Haidegger2022,Jorgensen2019}. However, most AI methods provide no measure of their prediction uncertainty, which makes it difficult to determine when the system is likely to make a prediction error. This poses a challenge when it comes to ensuring the safety of the system. In this paper, we aim to address this by presenting a framework for safe uncertainty-aware learning of autonomous robotic suturing.  

To better categorize the current efforts of autonomous RMIS, in \cite{attanasio2021autonomy}, Attanasio et al. define five distinct levels of autonomy in surgery. These levels provide a framework for categorizing the capabilities of autonomous surgical systems, focusing on the complexity of tasks it can perform, the system's independence, and the extent to which the system requires a surgeon's supervision.

Currently, most efforts in the literature consider the lower levels of autonomy, e.g., Level 1--3, where a system provides assistance to autonomous task execution with a surgeon's supervision. Most of these works consider small atomic medical tasks such as needle manipulation (e.g., grasping \cite{Liu2015a}, insertion, handling), suturing\cite{8710194,Iyer2013,Sen2016}, cutting, debridement, tissue piercing \cite{Zhong2019,Pedram20,Staub2010a,Jackson2013}, and knot-tying. These tasks are considered due to their high frequency in surgery, meaning automating them has the highest potential benefit. 
In particular, suturing is a crucial component of nearly all surgical procedures. Despite its significance, it remains a time-consuming and repetitive task that presents challenges when performed with surgical robots due to limited vision and/or haptic feedback, meaning that automating this task could significantly reduce surgeon fatigue and shorten operation times.

To automate these tasks, early methods relied on highly parametrized motions and specific starting conditions, and they also modified hardware or tools \cite{Sen2016,Zhong2019,Pedram20}, making them unsuitable for clinical use. 
Recent artificial intelligence (AI) advancements, particularly deep learning, are transforming autonomous surgery. Paradigms such as Learning from Demonstration (LfD) -- where the system learns from recordings of how an expert surgeon completes a task -- have gained popularity. Methodologies such as visuomotor policies \cite{chi2023diffusionpolicy} have shown promise for robotic autonomy in RMIS by effectively utilizing visual feedback to adapt to different task conditions and generalizations \cite{kim2024surgicalrobottransformersrt}.

Despite these advancements, medical professionals remain skeptical about using AI systems \cite{HUANG2024103223,wang2025aleatoricepistemicexploringuncertainty,Alowais2023,MOGLIA2021106151}. This skepticism primarily stems from two critical issues: these systems' inherent lack of explainability and the absence of robust safety guarantees.
Deep learning-based machine learning models often operate as 'black boxes,' making it difficult to understand how they make specific decisions. This opacity is problematic in the medical field, where the rationale behind clinical decisions must be transparent and comprehensible to ensure trust and accountability. Additionally, most approaches do not provide a measure of confidence in their predictions, or, if they do so, it is often poorly calibrated, showing little difference in the uncertainty reported for states the system has seen in the training data distribution (I.D.), and those that are Out-Of-Distribution (O.O.D.) -- i.e., that is has not been exposed to before. This makes it difficult to provide any form of safety guarantees for the predicted actions. The system could perform dangerous, unexplainable actions, subsequently furthering the surgeons' in such systems. 

Thus, if we wish to leverage the advantages of AI to design autonomous control systems for RMIS, quantifying model prediction uncertainty and providing safety guarantees in the control system's actions is critical. Autonomous RMIS systems must operate with high precision and reliability to prevent adverse outcomes. Additionally, the complexity of human anatomy and the variability in patient conditions demand that these systems can adapt to unforeseen scenarios without compromising safety. 

\begin{figure*}[ht!]
    \centering
    \includegraphics[width=\linewidth]{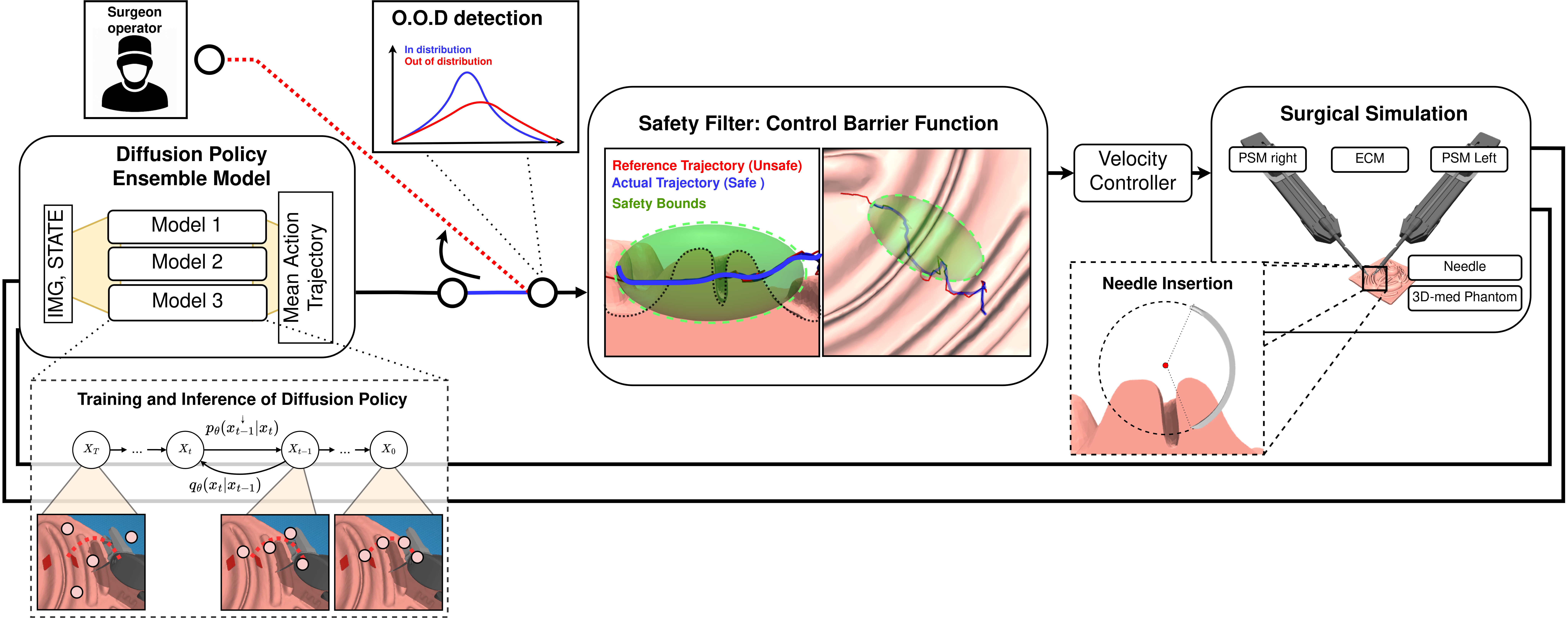}
    \caption{Our proposed framework consists of 1) a Diffusion Policy Ensemble model, which utilizes multiple trained Diffusion Policy models for quantifying epistemic uncertainty in the learned policy, 2) an O.O.D. detection scheme that is used to release control back to the surgeon, depeding on the uncertainty in the policy. It is possible to detect O.O.D. uncertainties based on a calibration test using this uncertainty measure, and  3) A Control Barrier Function, which acts as a safety filter on the robot controller and adds redundancy to safety measures by enforcing manually placed safety boundaries.}
    \label{fig:system_diagram}
\end{figure*}

This paper aims to answer the question: \textit{How can recent AI advances in policy learning be applied to autonomous robotic surgery, while ensuring that the robot actions remain safe even in situations that the system has not seen before?}

To do so, we present a framework for learning autonomous RMIS control policies with two redundant levels of: On the one hand, by using a diffusion policy ensemble to quantify prediction uncertainty, we are able to automatically detect when the model is operating I.D. and O.O.D., the latter of which could potentially result in unsafe actions. By detecting this, we are able to automatically relay control of the system to the surgeon when risk is high, in a way that corresponds with a Level 2 - 3 autonomous surgical system \cite{attanasio2021autonomy}. On the other hand, by applying model-free Control Barrier Functions to the robot's velocity controller, we can ensure that the output sent to the robot controller will stay within a defined safety region, even if our O.O.D. detection scheme fails. Note that this second safety mechanism can also be active when the surgeon is manually controlling the robot. An overview of our framework is provided in Fig.~\ref{fig:system_diagram}.

We evaluate our proposed approach in extensive experiments using a state-of-the-art robotic suturing simulator. We show that, using our approach, the learned policies can generalize to unseen conditions as long as they fall within the training data distribution, while potentially risky O.O.D. states can be automatically detected. Physically unsafe actions, such as pulling the needle out of the tissue and tearing the wound, are prevented by the safe CBF-based controller.

The remainder of the paper is structured as follows: \cref{sec:related_work} discusses related work, \cref{sec:preliminaries} presents the theoretical background for our method, and \cref{sec:methods} explains our methodology and our proposed framework. \Cref{sec:results} provides extensive evaluations of our approach in a surgical simulation environment, which we discuss in \cref{sec:discussion}, where we also analyze some of the limitations of our approach. Finally, we conclude the paper in \cref{sec:conclusion}.

\section{Related Work}
\label{sec:related_work}

This section will review related work on autonomy in surgical robotics, end-to-end learning of robot visuomotor policies, uncertainty quantification in machine learning, and formal safety guarantees for control systems using Control Barrier Functions.

\subsection{Autonomy in Robotic Surgery}

\subsubsection{Parametrization- and Planning-based Methods}
Many methods for autonomous robotic suturing have utilized planning and parametrization-based approaches \cite{Staub2010a,Jackson2013,Iyer2013,Liu2015a,Sen2016,Zhong2019,Pedram20}. A major limitation of these methods is their complexity: while the right parameter selection can result in highly optimized motions, selecting these parameters requires some degree of system understanding, which clinicians often lack. This can, to some extent, be addressed with additional systems and interfaces, such as in \cite{star2014} and \cite{8794306}, which allow surgeons to select and track incisions and stitch placements. Parameter selection can also be performed automatically, but this requires additional sensory input, typically computer vision, and can involve detecting or estimating the pose of objects in surgery, for instance, the needle or needle entry points during a suturing task \cite{Pedram20}. 
A limiting factor is that, under the wrong parameters, generalization to new task conditions, e.g.,  different suture points or needle placements, is severely limited and will usually lead to task failure. 

To make the systems more robust, some methods use modified needles or instruments \cite{Sen2016,Zhong2019,Pedram20} or fabricate specialized tools \cite{Leonard2014,Saeidi2019}, but this can render the methods unusable in a realistic clinical context, where the allowed equipment is highly regulated.

\subsubsection{Learning-based Methods}
To mitigate or avoid the need for parameter selection, researchers have explored learning-based methods for autonomous surgery. Multiple approaches \cite{Hutchinson_2023,7139344,schwaner2021autonomous} have studied the use of motion primitives learned from human demonstrations for automating medical sub-tasks. These methods typically combine learning at the skill level with segmentation and parametrization of motions, and hand-engineered finite state machines at the sub-task level \cite{7139344}.

A key direction involves transferring surgical skills from human demonstrations to robotic systems. \cite{9316273} employs Learning from Demonstration (LfD) to adapt open-surgery motions to Robot-Assisted Minimally Invasive Surgery (RAMIS); similarly, \cite{berg2010} leverages apprenticeship learning to achieve autonomous robotic execution with superhuman trajectory smoothness and speed performance. Further extending skill transfer, \cite{knoll2012} introduces scaffolded learning to enable human-machine collaboration in multi-arm robotic systems.

\subsubsection{Human-Robot Collaborative Frameworks}

Beyond autonomous execution, human-robot collaborative frameworks have been developed that can switch between manual and automated control dynamically. Padoy et al. proposed a system that seamlessly transitions between surgeon-guided operation and autonomous sub-task execution \cite{Padoy2011}. Expanding on this, Berthet-Rayne et al. introduced a structured approach combining active constraints, machine learning, and automated movements to enhance collaboration in bimanual tasks \cite{berthet-rayne2016}.

\subsection{End-to-end Robot Visuomotor Policies}

With recent advances in artificial intelligence (AI) models, various works have investigated end-to-end learning using neural networks. So-called visuomotor policies are control strategies that directly map image inputs to robot control command outputs. 
Florence et al. proposed Implicit Behavioral Cloning \cite{florence2022implicit}, based on an energy-based model architecture capable of representing discontinuous and multi-modal action distributions while exhibiting strong generalization capabilities. Diffusion Policy \cite{chi2023diffusionpolicy}, based on Denoising-Diffusion Probabilistic Models (DDPM) \cite{ho2020denoising}, demonstrates improved training stability compared to previous approaches, while retaining action multi-modality and strong generalization properties. Recently, the Transformer architecture was applied to surgical suturing, tissue retraction, and knot-tying tasks, demonstrating robustness to robot kinematic errors and generalization to unseen scenarios \cite{kim2024surgicalrobottransformersrt}.

\subsection{Uncertainty Quantification}
While end-to-end neural networks have recently shown great generalization potential, many models lack a measure of their prediction uncertainty or produce overconfident predictions. Even generative models, which theoretically model the data distribution, have been empirically shown to underestimate their prediction uncertainty, especially for O.O.D. states. Consequently, studies report mistrust in AI due to models' inexplicability, lack of safety guarantees, and absence of confidence measures in the learned policies \cite{HUANG2024103223,wang2025aleatoricepistemicexploringuncertainty,Alowais2023,MOGLIA2021106151}. This is particularly critical in safety-critical medical applications, where erroneous predictions can have severe clinical consequences. Therefore, well-calibrated uncertainty quantification is recognized as a crucial topic in advancing deep learning models across various domains
\cite{chen2022introductionexemplarsuncertaintydecomposition}.

In the following, we will review the main families of methods for uncertainty quantification in ML.

\subsubsection{Bayesian Inference methods}
One of the foremost methods of Bayesian inference-based methods for uncertainty quantification is Bayesian Neural Networks (BNNs), which assign probability distributions to all weights and biases. BNN makes it possible to decompose predictions into aleatoric and epistemic uncertainty, e.g., for classification tasks \cite{10.1007/978-3-030-86365-4_54} \cite{KWON2020106816}. A limitation of BNNs is that the process of training them is substantially more complicated and time-consuming than for traditional neural networks, often limiting their use in practice to smaller datasets and lower-dimensional problems.

Another branch of the literature utilizes Gaussian Processes (GP) \cite{williams2006gaussian}. Peng et al. \cite{peng2025uncertaintyunificationcasestudy} introduce a novel framework called Uncertainty-Unified Preference Learning (UUPL), which enhances GP-based preference learning by unifying human and robotic uncertainties. In \cite{xu2024onlinescalablegaussianprocesses}, GPs are integrated with conformal prediction to facilitate real-time uncertainty quantification, adapting thresholds based on feedback to ensure reliable coverage guarantees.

\subsubsection{Monte Carlo (MC) methods}
Monte Carlo Dropout (MCD) approximates Bayesian inference by applying dropout during training and testing. Multiple forward passes with dropout produce a distribution of predictions, offering an efficient way to estimate uncertainty without the high computational cost of methods such as deep Gaussian processes \cite{gal2016}. The variance of the predictions can be approximated as
\begin{equation} 
    \Sigma_y \approx \frac{1}{T}\sum_{t=1}^{T}(\hat{y}_t - \overline{y})^2, 
\end{equation}
where $T$ represents the number of forward passes, $\hat{y}_t$ denotes the prediction from the $t$-th forward pass, and $\overline{y}$ is the mean of the predictions.

\subsubsection{Model Ensemble methods}
Model ensemble methods train multiple independent neural networks to generate diverse predictions. The variability among these predictions encodes a proxy of the prediction uncertainty, as different models may learn different representations, capturing epistemic uncertainty.\\
Ensemble learning involves training $\mathcal{M}$ models with different initial conditions, e.g., weights or data. Ensemble learning
leverages the prediction entropy of having multiple models. Multiple predictions make it possible to calculate metrics such as the mean and variance as seen in Eq.~\ref{eq:ensemble_model}, where $\theta_i$ is the model parameters of model $i$.

\begin{equation}
    p(y|x) = \frac{1}{M}\sum^{M}_{i=1}p(y|x,\theta_i) 
    \label{eq:ensemble_model}
\end{equation}

Model ensemble has been used to enhance the reliability of autonomous surgical robots by improving their ability to detect task failure early \cite{thompson2025earlyfailuredetectionautonomous}, as well as for medical imaging tasks, such as segmentation \cite{chen2022introductionexemplarsuncertaintydecomposition}, \cite{thompson2025earlyfailuredetectionautonomous}. Due to labeled data being costly for medical imaging, robotics, and computer vision,  \cite{10648946} uses an ensemble of Gaussian processes for active learning.

Ensemble learning has empirically been found to produce better calibrated uncertainty estimates across most metrics than other methods \cite{ovadia2019can}; however, it does so at increased computational cost, memory requirements, training times, and inference speed.

\subsection{Control Barrier Functions}

Although safety in a learned control policy can be addressed from the ML side by quantifying prediction uncertainty, it can also be approached from the point of view of control theory.

In control theory, Control Barrier Functions (CBFs) \cite{ames2019control} are a well-established method to achieve formal safety guarantees by ensuring the forward set invariance and can be efficiently integrated with a quadratic programming  \cite{Ames2017CBFQP}. This has enabled CBFs to be incorporated in a wide range of modern robotics applications, such as laboratory automation \cite{WOLF202218}, force control \cite{dawson2022barrier}, input-delayed systems \cite{YKECC2024}, robotic grasping \cite{kim2024robustadaptivesaferobotic}, and medical robotics \cite{Frederik2024medicalforce}\cite{kim2025safetyensuredcontrolframeworkrobotic}. However, safety guarantees in the standard CBF formulation can be compromised due to their high dependency on system model accuracy. 

\subsubsection{Handling Modeling Uncertainty}

Multiple methods have been proposed to address the sensitivity of CBFs to model uncertainty. Robust CBFs \cite{JANKOVIC2018359} accomplish robust, but conservative safety by enforcing robustness against unknown disturbances in the safety constraints. Adaptive CBFs \cite{Taylor2020} use conditional parameter estimation for forward invariance of the safe set. However, these methods tend to be overly conservative, hampering control performance. To tackle this, Robust adaptive CBFs (RaCBFs) \cite{lopez2020robust} integrate both methods. Data-driven methods, such as Gaussian Processes (GP), can be used to estimate and compensate for unknown modeling dynamics by updating the posterior of the safety condition online \cite{castaneda2021gaussian}. The GP and Robust adaptive methods can also be combined, resulting in lower model uncertainty and, consequently, reduced conservatism \cite{kim2024safe}. CBFs-based safety controllers have also been integrated with disturbance observers to estimate and compensate for unknown model dynamics, while alleviating the conservatism of robust control schemes. Observer-based methods have considered state uncertainty \cite{Wang2022ACC_EEQ} and external disturbances \cite{Ersin2022}\cite{AlanDOBCBF2023}. 

\subsubsection{Model-Free Strategies}

In practice, modeling the full-order system dynamics is non-trivial, so model-free methods with formal safety guarantees have also been considered. Model-free barrier functions using data-driven techniques were first introduced to design safe controllers for complex system dynamics \cite{squires2022modelfreebarrierfunctions}. In \cite{Singletary2022}, they proposed a framework for collision avoidance that generates a safe velocity profile by combining artificial potential fields and CBFs, without the need to consider the complex dynamics of a quadrotor. The recently-introduced \textit{Model-free CBF} framework \cite{molnar2021model}, formalizes and provides the design of a safe velocity controller simply by assuming that a velocity-tracking controller is available. As this type of controller is already widely used in robotics \cite{Spong2005}, model-free CBFs can be efficiently applied to many applications involving complex actuator mechanisms, such as the tendon-driven robots used in RMIS.\\

\section{Preliminaries}
\label{sec:preliminaries}

\subsection{Imitation Learning}
Imitation learning (IL) is a type of machine learning in which an agent learns to perform tasks by mimicking the behavior of an expert. This approach is beneficial when it is difficult to define a reward function for reinforcement learning or when expert demonstrations are readily available. The simplest form of IL is Behavioral Cloning (BC), which learns a policy, $f_\theta$ that maps observations of the state, $o$, to actions, $a$, closely replicating the expert's behavior, i.e., $\hat{a}=f_\theta(o)$. This explicit model does not handle multi-modality, an important attribute for more complex tasks. Therefore, in \cite{florence2022implicit}, the authors introduce an energy-based BC, Implicit Behavioral Cloning (IBC), i.e., Implicit Model, which is formulated as \(\hat{a}=\arg \min_{a\in\mathcal{A}} E_\theta(o, a)\), where $E_\theta$ is a so-called \textit{energy function}, which is equivalent to an unnormalized probability distributions. These Energy Models handle multi-modality well, making them well-suited for our task.

\subsubsection{Diffusion Policy}
\begin{figure}[!h]
    \centering
    \includegraphics[width=\linewidth]{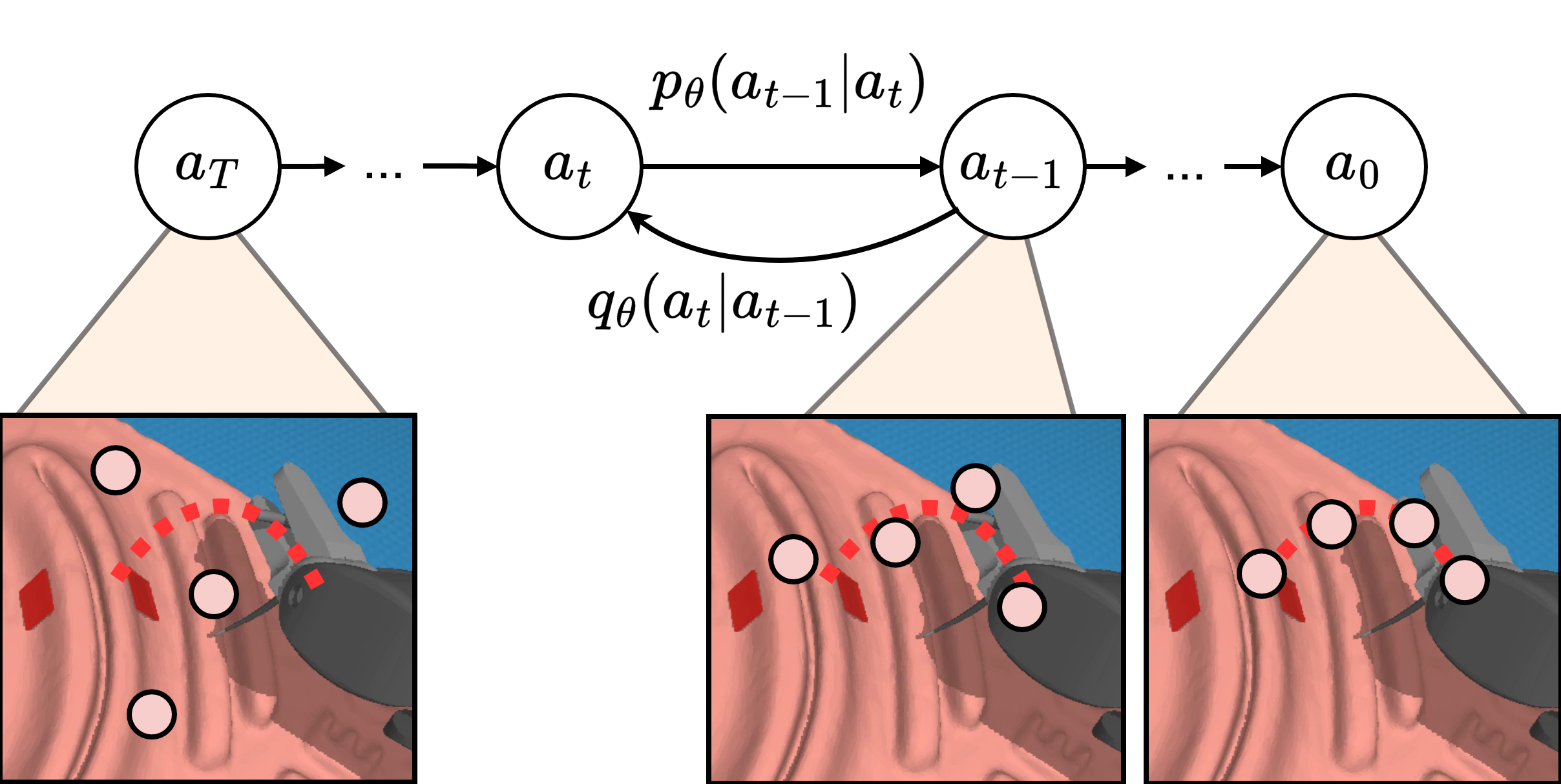}
    \caption{Illustrations of how the diffusion process is applied on the demonstration trajectories and how the model trains to reverse the diffusion process}
    \label{fig:diffusion_process}
\end{figure}

Building upon IBC, the authors in \cite{chi2023diffusionpolicy} introduce Diffusion Policy, which leverages Denoising Diffusion Probabilistic Models (DDPMs) \cite{ho2020denoising}. IBC and Diffusion Policy are closely related, as IBC models the energy landscape, and Diffusion Policy models the gradient of the energy landscape. DDPMs are generative models where the output is a denoising process. Starting from $a_K$ sampled from Gaussian noise, the DDPM performs $K$ iterations of denoising and produces a series of actions with decreasing noise, $a_k, a_{k-1}, \ldots, a_0$, until a desired noise-free action $a_0$ is formed. See Fig.~\ref{fig:diffusion_process}. The process follows the equation:

\begin{equation}
a_{k-1} = \alpha (a_k - \gamma \epsilon_\theta (o, a_k, k) + \mathcal{N}(0, \sigma^2 \mathbf{I})),
\label{eq:DPEQ1}
\end{equation}

where $\epsilon_\theta$ is the noise prediction network conditioned on the current observation and action, $\theta$ are its parameters, which are optimized through learning, and $\mathcal{N}(0, \sigma^2 \mathbf{I})$ represents Gaussian noise added at each iteration. Eq. \ref{eq:DPEQ1} can also be interpreted as a single noisy gradient descent step:

\begin{equation}
x' = x - \gamma \nabla E(x),
\end{equation}
where the noise prediction network $\epsilon_\theta(o, x, k)$ predicts the gradient field $\nabla E(x)$, with $\gamma$ being the learning rate.\\

\subsection{Likelihood Ratio testing}
Likelihood Ratio testing (LRT) is a statistical method useful for comparing two models. It compares which model better fits the observed data. 

\subsubsection{Likelihood Function}
The likelihood function is denoted \(L(\theta | x)\), which measures how well a model explains the observed data, i.e., \(x_1, x_2, ..., x_n\) given parameters \(\theta\). The \(f(x_i | \theta)\) is the PDF. 

\begin{equation}
    L(\theta | x) = \Pi _{i=0}^n f(x_i | \theta)
\end{equation}

\subsubsection{Likelihood-Ratio test statistic}
The likelihood-ratio statistic $\lambda$ is the ratio of \(L(\hat \theta_0 | x)\), i.e., the maximum likelihood under the null model, and \(L(\hat \theta | x)\), i.e., the maximum likelihood of the alternative model. 

\begin{equation}
    \lambda = \frac{L(\hat \theta_0 | x)}{L(\hat \theta | x)}
\end{equation}

\subsection{Model-Free Control Barrier Functions}
Consider the following full-order robot dynamics:
\begin{equation}
M(\bm{q}) \ddot{\bm{q}} + C(\bm{q}, \dot{\bm{q}}) \dot{\bm{q}} + g(\bm{q}) = B\bm{u}, \label{pre:robot_dynamics}
\end{equation}
where $\bm{q} \in \mathbb{R}^n$ are the system configurations in configuration space, $\mathrm{Q}$, and $ M(\bm{q}) \in \mathbb{R}^{n \times n}$ is the inertia matrix,  $ C(\bm{q}, \dot{\bm{q}}) \in \mathbb{R}^{n \times n}$ denotes the Coriolis and centrifugal terms, $g(\bm{q}) \in \mathbb{R}^n$ represents the gravity, and $B\in \mathbb{R}^{n\times m}$ is the input property matrix. A safe set with a continuously differentiable function, $h_q: \mathrm{Q} \rightarrow \mathbb{R}$ is defined as $\mathcal{S}_q = \{\bm{q} \in \mathrm{Q} \text{ }\vert\text{ } h_q(\bm{q}) \geq 0\}$, and we assume that $\big|\big|\frac{\partial h_q(\bm{q})}{\partial \bm{q}}\big|\big| \leq C_h$, $C_h>0$ for $\forall \bm{q} \in \mathcal{S}_q$ and an exponentially stable velocity controller, $\bm{u} = k_q(\bm{q},\dot{\bm{q}})$, $k_q:\mathrm{Q}\times \mathbb{R}^n \rightarrow \mathcal{U}$, is provided such that $||\dot{\bm{e}}(t)||\leq M||\dot{\bm{e}}_0||\mathrm{e}^{-\lambda t}$ for the non-negative scalar, $M, \lambda$. That means there exists a continuously differentiable Lyapunov function, $V(\bm{q},\dot{\bm{e}})$ such that $k_1||\dot{\bm{e}}|| \leq V(\bm{q},\dot{\bm{e}}) \leq k_2||\dot{\bm{e}}||$ and $\dot{V}(\bm{q}, \dot{\bm{e}},\dot{\bm{q}}, \ddot{\bm{q}}_d, \bm{u}) \leq -\lambda V(\bm{q},\dot{\bm{e}})$ where $k_1, k_2 > 0$. Note that $\dot{\bm{e}} = \dot{\bm{q}} - \dot{\bm{q}}_d$ is the velocity tracking error from the desired velocity, $\dot{\bm{q}}_d \in \mathbb{R}^n$. Consequently, the model-free safety controller ensures safety guarantees based on the following theorem:
\begin{theorem}[\cite{molnar2021model}]
    Consider the safe set, $\mathcal{S}_q$ with respect to \eqref{pre:robot_dynamics}, and safe velocity, $\dot{\bm{q}}_s$ satisfying 
        \begin{equation}
        \frac{\partial h_q(\bm{q})}{\partial \bm{q}}\dot{\bm{q}}_s \geq -\alpha\big(h_q(\bm{q})\big), \quad \forall\bm{q}\in \mathcal{S}_q. \label{pre:model_free_safety_condition}
    \end{equation}
    The system \eqref{pre:robot_dynamics} is safe with respect to $\mathcal{S}_q$ if the velocity controller, $\bm{u} = k_q(\bm{q},\dot{\bm{q}})$ with $\lambda > \alpha$ tracks a safe velocity, $\dot{\bm{q}}_s$ such that the initial condition, $(\bm{q}_0,\dot{\bm{e}}_0) \in \mathcal{S}_V$ where $\mathcal{S}_V = \{(\bm{q},\dot{\bm{e}}) \in \mathrm{Q} \times \mathbb{R}^n \text{ }:\text{ } h_V(\bm{q},\dot{\bm{e}}) = -V(\bm{q},\dot{\bm{e}}) + \alpha_e h_q(\bm{q}) \geq 0\}$ with $\alpha_e = \frac{k_1(\lambda -\alpha)}{C_h}$.

    \label{pre:theorem_mfcbf}
\end{theorem} 

External disturbances, $\bm{d}\in\mathbb{R}^m$, but bounded $||\bm{d}||_{\infty} \leq k_3 \in \mathbb{R}_{>0}$, often happen in practice and can be incorporated as follows:
\begin{equation}
    M(\bm{q}) \ddot{\bm{q}} + C(\bm{q}, \dot{\bm{q}}) \dot{\bm{q}} + g(\bm{q}) = B(\bm{u}+\bm{d}). \label{pre:robot_dynamics_dis}
\end{equation}
To tackle this challenge, a robust model-free safe controller can be designed in the following corollary.
\begin{corollary}[\cite{molnar2021model}]
Consider inflated safe sets, $\mathcal{S}_d \supseteq \mathcal{S}_q$, $\mathcal{S}_d = \{\bm{q} \in \mathrm{Q} \text{ }\vert\text{ }h_d(\bm{q})\geq 0\}$ where $ h_d(\bm{q}) = h_q(\bm{q}) + \gamma(||\bm{d}||_{\infty})$ and $\mathcal{S}_{V_d} = \{(\bm{q},\dot{\bm{e}}) \in \mathrm{Q} \times \mathbb{R}^n \text{ }:\text{ } h_{V_d}(\bm{q},\dot{\bm{e}}) \geq 0\}$ where $h_{V_d}(\bm{q},\dot{\bm{e}}) = h_V(\bm{q},\dot{\bm{e}}) +\gamma(||\bm{d}||_{\infty})$, and $\gamma$ is class $\mathcal{K}$ function. And let us assume that there exists a robust stable velocity tracking controller such that \textit{Input-to-State exponential Stability} (ISS) \cite{sontagISS2008} and  the initial condition, $(\bm{q}_0,\dot{\bm{e}}_0)\in \mathcal{S}_{V_d}$:
    \begin{equation}
        \dot{V}(\bm{q}, \dot{\bm{e}},\dot{\bm{q}}, \ddot{\bm{q}}_d, \bm{u},\bm{d}) \leq -\lambda V(\bm{q},\dot{\bm{e}}) +\iota(||\bm{d}||_{\infty}), \quad \lambda > \alpha, \label{pre:ISS_vel_ctrl}
    \end{equation}
    where $\iota$ denotes an extended class $\mathcal{K}^{e}_{\infty}$ function.
    If the velocity controller \eqref{pre:ISS_vel_ctrl} tracks $\dot{\bm{q}}_s$ ensuring $ \frac{\partial h_d(\bm{q})}{\partial \bm{q}}\dot{\bm{q}}_s \geq -\alpha\big(h_d(\bm{q})\big), \forall\bm{q}\in \mathcal{S}_d$, the system \eqref{pre:robot_dynamics_dis} is \textit{Input-to-State Safety} (ISSf) \cite{ShishirISSfCBFs2019} with respect to $\mathcal{S}_d$.
    \label{pre:ISSf-model_free_cbf}
\end{corollary}

\section{Methods}
\label{sec:methods}
In this section, we describe our framework for ensuring the safety of learned robot control policies at two stages of the control pipeline. Based on expert demonstrations of suturing tasks, we train an ensemble of Diffusion Policy agents. Each Diffusion Policy model in the ensemble takes in the internal states of the surgical robot, i.e., the joint position of the Patient-Side Manipulator (PSM), as well as the image captured by the Endoscopic Camera Manipulator (ECM).
With this data, each of $N$ models that make up the Diffusion Policy Ensemble predicts an action sequence. By considering the (dis-)agreement between each action sequence in the ensemble, we can quantify prediction uncertainty and use a simple likelihood ratio test to classify whether the model is operating I.D. or O.O.D.. In the O.O.D. case, the state is deemed anomalous -- therefore, potentially unsafe -- and control is given back to the surgeon. As a redundant safety mechanism, we use control barrier functions to place formal safety guarantees in critical areas of the state-space (for example, near the suturing points), keeping the control action sent to the robot velocity controller safe even in the event of unsafe policy predictions not caught by our O.O.D. detection. An overview of our framework can be seen in \Cref{fig:system_diagram}. The following sections will describe each system component.

\subsection{Data Collection}
We use a simulation environment consisting of two PSM arms, one ECM arm, one 3D-Med phantom, and a Needle. The simulation is the framework presented in \cite{munawar2022}, building upon the Asynchronous Multi-Body Framework (AMBF) introduced initially for the AccelNet Surgical Robotics Challenge 2023-2024. With this simulator, it is possible to generate synthetic training data. 

\begin{figure}[h]
    \centering
    \includegraphics[width=\linewidth]{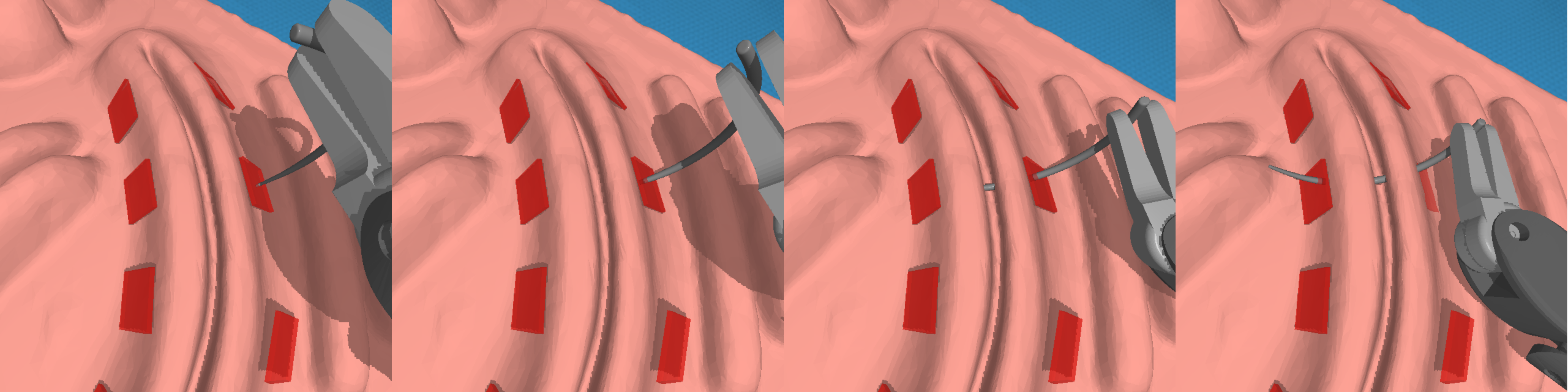}
    \caption{Image sequence showing a needle insertion demonstration.}
    \label{fig:task_insertion}
\end{figure}

The current setup with a suturing phantom and needle is ideal for considering the needle insertion task. Fig.~\ref{fig:task_insertion} depicts the needle insertion task. The needle is initialized and placed in the right PSM gripper. Having the needle already in the PSM gripper also resembles the real usage of a surgeon activating the autonomous system for a takeover during a procedure. The 3D-Med Phantom is moved during each instance, so the system must move the needle tip toward the suture entry point using visual servoing and then, perform a needle insertion. 

To generate synthetic data autonomously, we defined an expert demonstrator by using privileged data, i.e., data not seen by our Diffusion Policy Model, from the simulator, such as the ground truth needle position and suture entry point, to generate perfect needle insertion motions. We generate $\sim$100 demonstrations for training and $\sim$33 for validation. We formulate a single state as $(x, y, z) \in \mathbb{R}^3$, the first two columns of a rotation matrix $O \in \mathrm{SO}(3)$, and the gripper state (Only considering one PSM).

\subsection{Training Diffusion Policies}
We implement a Diffusion Policy as specified in \cite{chi2023diffusionpolicy}. Diffusion Policy relies on a Denoising Model, essentially turning pure noise into a learned action distribution based on conditional observations. See \cite{ho2020denoising} for further information on this process. To achieve this, we sample a unmodified $a_{t,0}$ trajectory from the set of demonstrations, which is split into action segments of length $T_p$ and corresponding conditioning observations of length $T_o$ to train this model. Then, iteratively, $k$ iterations more noise $\epsilon_k$ are added to the ground truth action segments, and the denoising model is trained\footnote{All models were trained on an NVIDIA GeForce RTX 3090 GPU.} to predict the added noise using the loss function described in Eq.~\ref{eq:DPLMSE}. Our model hyperparameters are shown in Table \ref{tab:hyperparameters}.
\begin{equation}
\mathcal{L} = \text{MSE}\left(\epsilon_k, \epsilon_\theta\left(a_{t,0} + \epsilon_k, k\right)\right).
\label{eq:DPLMSE}
\end{equation}

By learning to predict the noise, it is possible to reverse the process, i.e., start with pure noise samples and remove the noise iteratively until an action segment is reached, as described in Eq.~\ref{eq:DPstep}.

\begin{equation}
x_{k-1} = \alpha (x_k - \gamma \epsilon_\theta (x_k, k) + \mathcal{N}(0, \sigma^2 \mathbf{I})),
\label{eq:DPstep}
\end{equation}

To ensure temporal consistency of the actions -- often a problem with other BC methods -- $T_p$ time steps into the future are predicted, but only $T_a$ are executed, with $T_a < T_p$. Similar methods are used in Model Predictive Control.

\begin{table}[h!]
    \centering
    \caption{Hyperparameters for the Diffusion Policy Model.}
    \label{tab:hyperparameters}
    \begin{tabular}{ll|ll}
        \textbf{Hyperparameter} & \textbf{Value} & \textbf{Hyperparameter} & \textbf{Value} \\
        prediction horizon    & 8  & batch size             & 8 \\
        observation horizon     & 4   & num epochs             & 10000 \\
        action horizon  & 4   & lr (learning rate)      & $1 \times 10^{-4}$ \\
        observation dim         & 10  & weight decay           & $0.1$ \\
        action dim      & 10  & number of warmup steps      & 100 \\
                          &     & number of diffusion iters.  & 100 \\
    \end{tabular}
\end{table}

\begin{figure}[h]
    \centering
    \includegraphics[width=\linewidth]{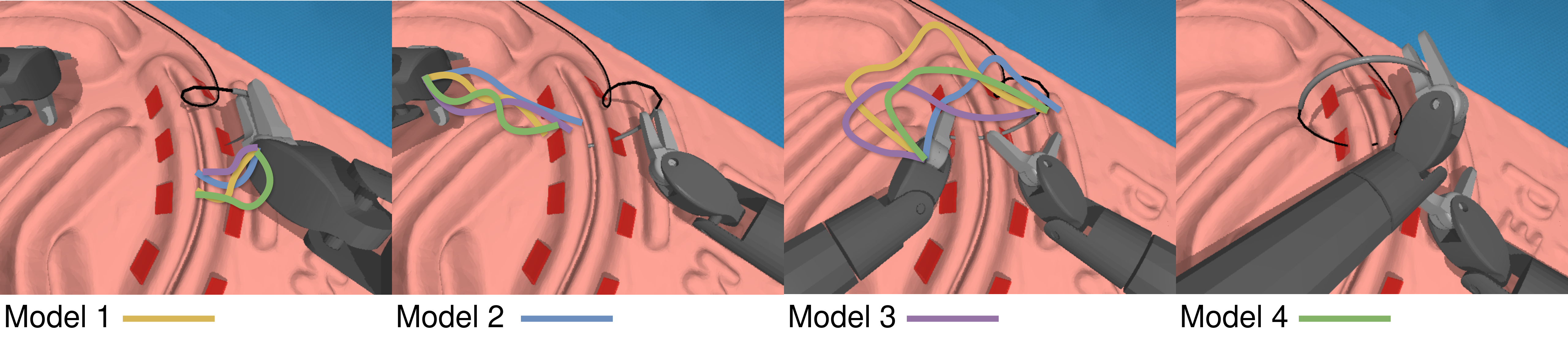}
    \caption{Image sequence of a single suture throw execution with a model ensemble consisting of four trained diffusion policy models' predictions visualized.}
    \label{fig:model_ensemble}
\end{figure}

\subsection{Ensemble Model Predictions}
We create a Model Ensemble by training $N$ Diffusion Policy models on the same training data. Intuitively, when the models are trained independently, the variation in their parameters will lead to different predictions for each model in the ensemble and thereby a distribution over predictions. For scenarios in the training data distribution, the variance will be low. However, during O.O.D. scenarios, the predictions will vary significantly, allowing for quantification of the model's epistemic uncertainty.

At each inference step, we predict $N$ \textit{single-model action sequences}, one for each model in the ensemble, and each spanning a horizon $T_P$, i.e., $\mathbf{a}_n = \left[a_{n, t}, a_{n, t+1}, \dots, a_{n, t+T_P} \right]$ for $n=1, \dots, N$, and aggregate these into an \textit{ensemble action sequence} matrix $\mathbf{A} = [\mathbf{a}_1, \mathbf{a}_{2}, \dots, \mathbf{a}_{N}]^{\top}$, as can be seen in Fig.~\ref{fig:model_ensemble}. For each time step in the \textit{ensemble action sequence} $\mathbf{A}$, we aggregate the ensemble predictions into a single \textit{mean action sequence}, $\mathbf{\mu}_a = \left[\mu_{a, t}, \mu_{a, t+1}, \dots, \mu_{a, t+T_P} \right]$, which is used to control the robot. We also calculate a \textit{covariance of the action sequence}, $\mathbf{\Sigma}_a = \left[\Sigma_{a, t}, \Sigma_{a, t+1}, \dots, \Sigma_{a, t+T_P} \right]$, which is used for uncertainty quantification and O.O.D. detection, as will be detailed in the next section.

\subsection{Out-of-Distribution Detection}
We use the maximum value of the absolute action covariance, $\hat{\sigma}_t = \max \lvert \Sigma_{a,t} \rvert$, at each step in an action sequence to quantify prediction uncertainty.

To detect when the model is operating O.O.D., and is thus likely to produce erroneous predictions, we leverage a likelihood ratio test (LRT). To utilize this test, an I.D. model needs to be created $\mathcal{M}_{ID} = \mathcal{N}(\bar x,h\sigma)$, which requires a calibration process where the model executes the policy within I.D. scenarios, and the corresponding prediction uncertainty is measured. During inference, after an ensemble prediction step as outlined in the previous section, and before an action is executed on the robot controller, we determine whether the sample is I.D. or O.O.D. using an LRT with a significance factor of $\alpha=0.05$. If a sample is O.O.D., the current state is deemed anomalous (at least with respect to the training data) and the system releases control back to the surgeon.

\subsection{Safety Guarantees on the Robot Controller}
Abrupt and unsafe motions should be prevented in advance during robotic suturing automation since they might lead to serious damage to normal tissues. In our case, this stage adds a redundant safety layer to the O.O.D. scheme presented above; even if O.O.D. fails, the controlled robot action should be made safe. To this end, we follow the previous work from \cite{molnar2021model} to incorporate the model-free CBFs-based safety filter \eqref{pre:model_free_safety_condition} into the proposed framework. This ensures that the robot's movements are constrained to remain within a safe region, which activates once the robot begins a suture throw and which prevents the robot from, e.g., pulling the needle violently in a way that would potentially tear the tissue. The safe region is defined as an ellipsoid with the length of semi-axes, $a,b,c$, and the origin, $P_o\in \mathbb{R}^3$. Subsequently, we define a candidate control barrier function based on the ellipsoid in the following: 
\begin{equation}
h(\bm{x}_c) = 1-\Bigg(\frac{D_x^2}{a^2}+\frac{D_y^2}{b^2}+\frac{D_z^2}{c^2}\Bigg) , \label{exp:a_tumor_cbf}
\end{equation}
where $D =  \bm{x}_c-P_o$ is the vector containing elements $\{D_x,D_y,D_z\}^{\top}$ and $\bm{x}_c\in \mathbb{R}^3$ is the end-effector position of the robot. Consequently, the safety constraint is defined as:
\begin{equation}
\nabla h(\bm{x}_c)\dot{\bm{x}}_s \geq -\alpha h(\bm{x}_c) , \label{exp:cbf_cond}
\end{equation}
where $\dot{\bm{x}}_s \in \mathbb{R}^3$ is the safe velocity and $\nabla h(\bm{x}_c) = -\big\{\frac{D_x}{a^2}, \frac{D_y}{b^2}, \frac{D_z}{c^2}\big\}/||\{\frac{D_x}{a^2}, \frac{D_y}{b^2}, \frac{D_z}{c^2}\big\}|| $ is the derivative of the safe set. Furthermore, $\nabla h(\bm{x}_c)^{\top}$ can be denoted by the directional vector, $\mathsf{n}_o\in \mathbb{R}^3$, between the robot and the origin of the ellipsoid. To obtain the a velocity satisfying \eqref{exp:cbf_cond}, we use the following solution by \cite{molnar2021model}:
\begin{equation}
    \dot{\bm{x}}_s = \dot{\bm{x}}_n + \max\big(-\mathsf{n}_o^{\top}\dot{\bm{x}}_n-\alpha(h(\bm{x}_c)),0\big)\mathsf{n}_o,
\end{equation}
where $\dot{\bm{x}}_n\in \mathbb{R}^3 $ is the nominal velocity control input. If the stable velocity controller tracks the safe velocity, $\dot{\bm{x}}_s$, it is ensured that the robot stays within the user-defined safe region, thus accomplishing the formal safety guarantees given by \textit{Theorem~\ref{pre:theorem_mfcbf}}.

\section{Results}
\label{sec:results}

In this section, we present the results of several experiments that aim to highlight the efficacy of our framework. 
First, we demonstrate the system's ability to quantify epistemic uncertainty using the learned Diffusion policy models. Concurrently, we use O.O.D. detection on the estimated uncertainty to classify I.D. or O.O.D. scenarios. We test several O.O.D scenarios, e.g., dropping the needle, moving the camera, and unexpectedly moving the phantom.
Note that, for all of these experiments, O.O.D. detection is shown but not enforced, i.e., even when O.O.D. is successfully detected, the policy is allowed to continue to more easily visualize the resulting behavior.
In an additional experiment, we show that, using model-free CBFs, unsafe control actions that would violate the defined safety bounds can be filtered and made safe, in this case, by avoiding pulling the needle outside of the wound once a suture throw is started.

\subsection{Experiment 1: Needle dropped during execution}
\begin{figure}[h]
    \centering
    \includegraphics[width=\linewidth]{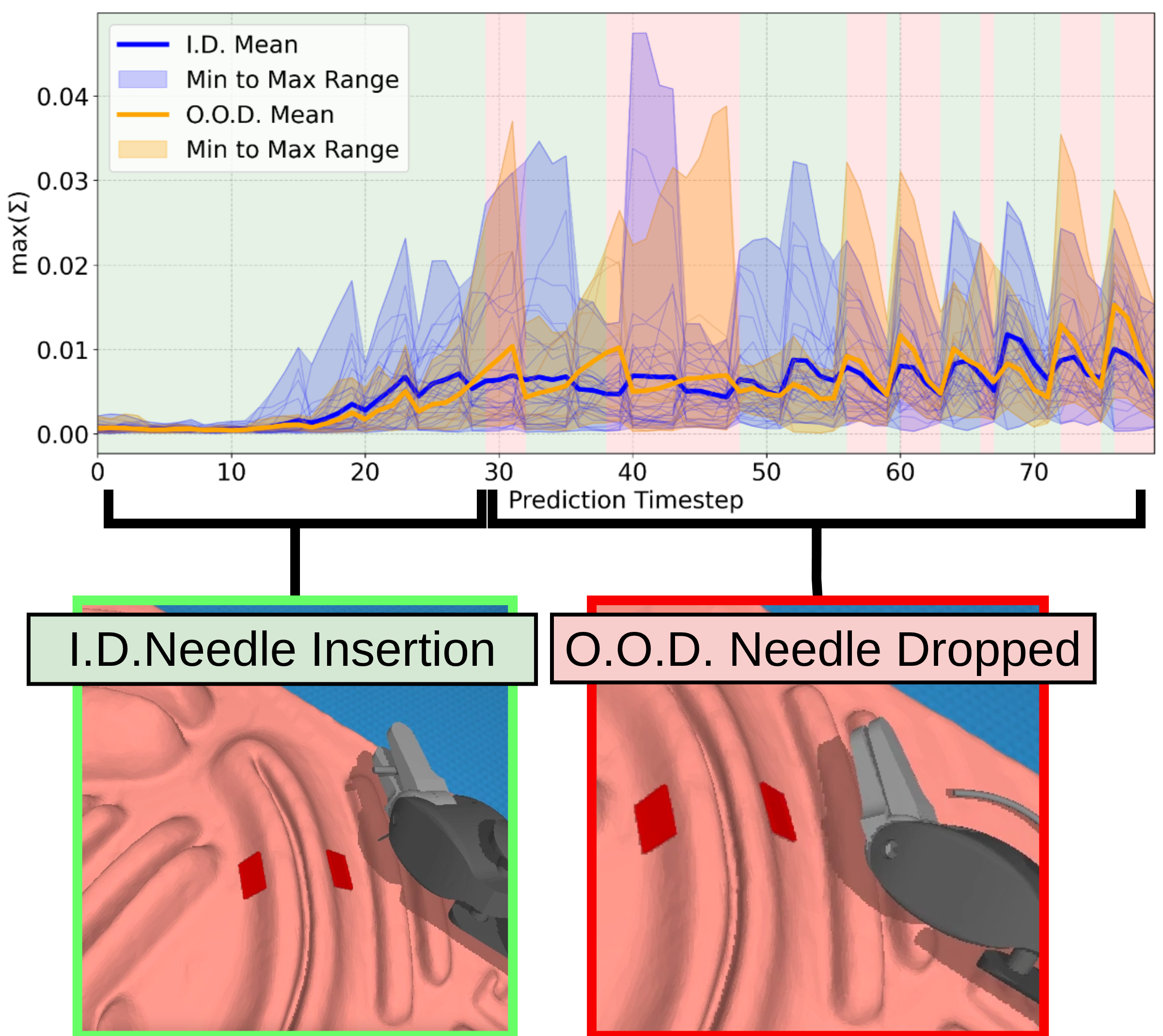}
    \caption{The top graph depicts the estimated uncertainty ($\max (\Sigma)$) for each prediction at each timestep by the model ensemble. The background indicates the prediction of the O.O.D. detector, i.e., green for I.D. and red for O.O.D.. We execute the scenario 12 times. We use the maximum sample of the 12 runs during the O.O.D detection. The bottom images show that the scenario starts with I.D. needle insertion; however, at timestep 8, the needle is dropped and detected at timestep 30.}
    \label{fig:UQ_NeedleDropped}
\end{figure}

In this experiment, we examine the scenario where the needle is dropped during executions, a realistic case during surgical procedures. 
The estimated uncertainty is illustrated in Fig.~\ref{fig:UQ_NeedleDropped}. The model successfully detects the dropped needle throughout multiple executions. The results of the O.O.D detector are depicted as the colored background, where red is O.O.D and green is I.D.

\subsection{Experiment 2: Moved Camera}
In this experiment, we examine the scenario where the camera is moved within the scene. While camera movement is common in surgery, this experiment simulates a situation where the camera is positioned sub-optimally at a poor viewing angle, which the model has never seen before in training.

\begin{figure}[h]
    \centering
    \includegraphics[width=\linewidth]{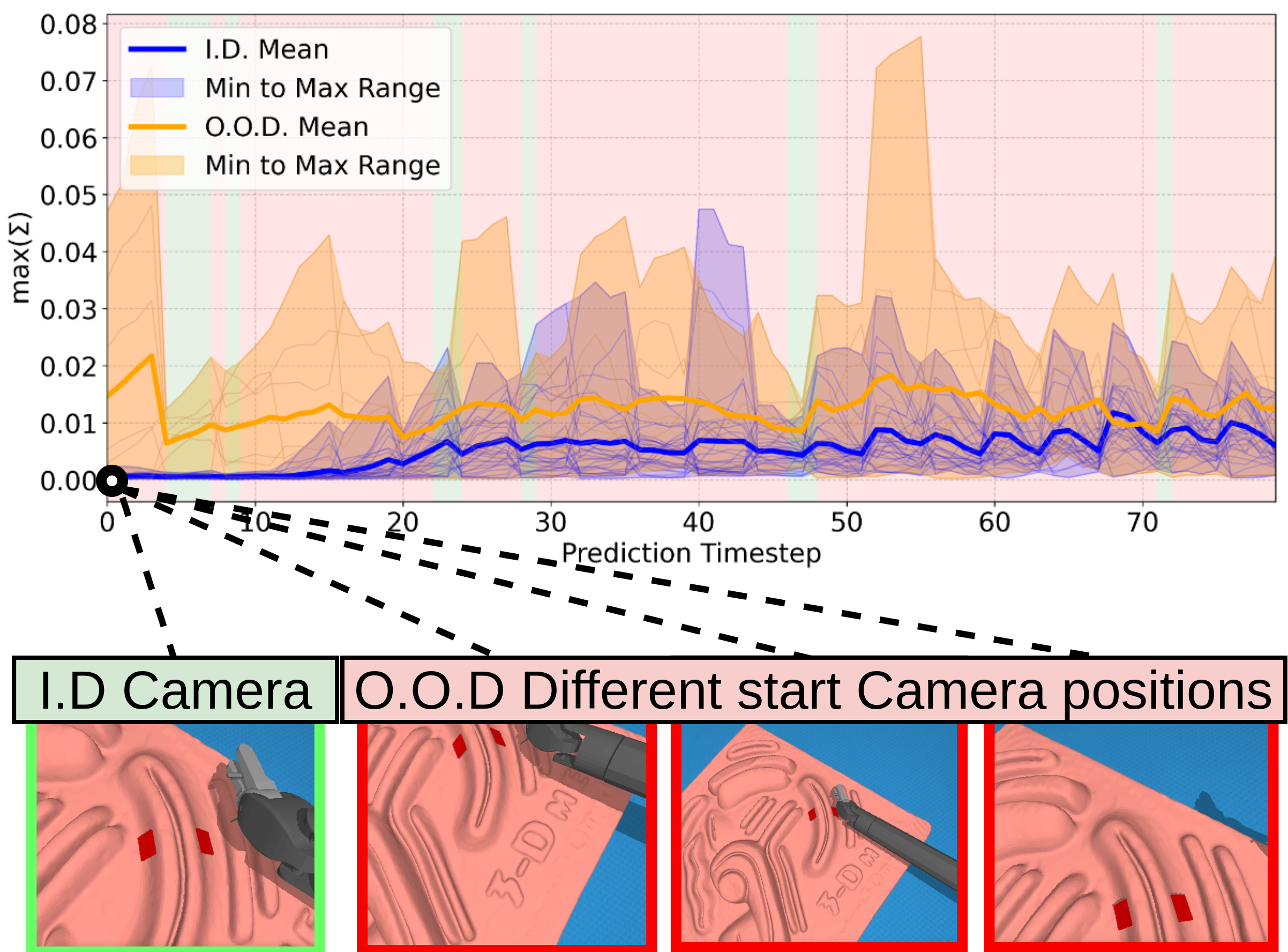}
    \caption{The top graph depicts the estimated uncertainty ($\max (\Sigma)$) for each prediction at each timestep by the model ensemble. The background indicates the prediction of the O.O.D. detector, i.e., green for I.D. and red for O.O.D.. We execute the scenario 8 times. We use the maximum sample of the 8 runs during the O.O.D detection. The bottom images show that in this scenario, the camera viewing angle is changed for each execution, which is detected throughout the executions as O.O.D. 
   }
    \label{fig:UQ_MovedCamera}
\end{figure}

The estimated uncertainty is illustrated in Fig.~\ref{fig:UQ_MovedCamera}. It is evident that the suboptimal camera positions significantly impact the prediction uncertainty compared to the baseline. In this scenario, the O.O.D. detection is highly active as the visual discrepancy between the I.D. and the O.O.D. is large.

\subsection{Experiment 3: Moving Phantom}
In this experiment, we investigate the scenario of moving the phantom within the scene. Sudden movements due to organ or tissue deformations are common in surgery. We simulate this by repositioning the phantom to various locations, demonstrating the servoing behavior of visuomotor policies. Additionally, we move the phantom after insertion to showcase the policy's robustness.
We perform two scenarios:

\subsubsection{Experiment 3.1: Servoing}
\begin{figure}[h]
    \centering
    \includegraphics[width=\linewidth]{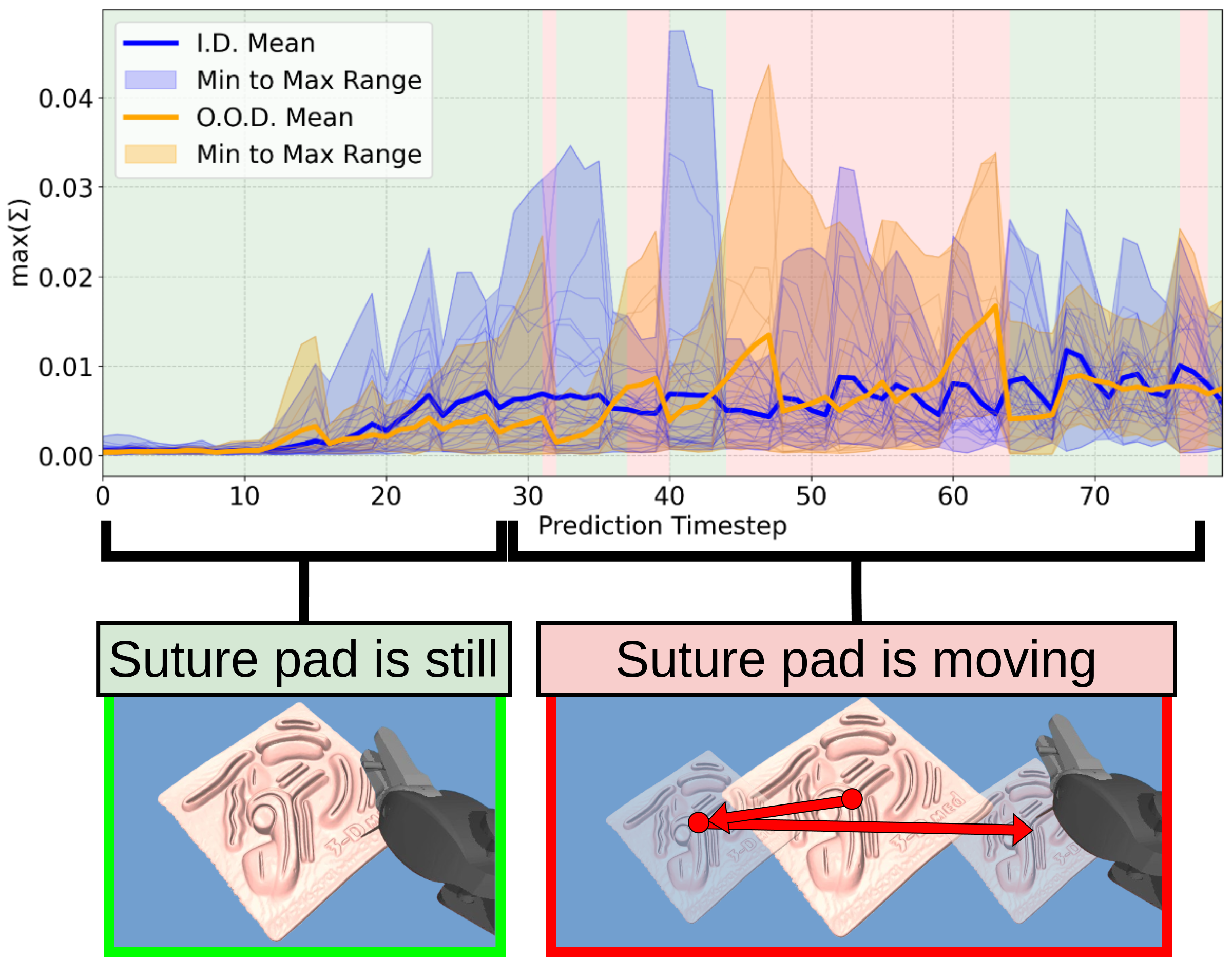}
    \caption{The top graph depicts the estimated uncertainty ($\max (\Sigma)$) for each prediction at each timestep by the model ensemble. The background indicates the prediction of the O.O.D. detector, i.e., green for I.D. and red for O.O.D.. We execute the scenario 9 times. We use the maximum sample of the nine runs during the O.O.D detection. The bottom images show that at the beginning of the scenario, the suture pad remains still; however, it changes position periodically.}
    \label{fig:UQ_Servo}
\end{figure}

The estimated uncertainty is illustrated in Fig.~\ref{fig:UQ_Servo}. Moving the phantom during execution does not always increase the uncertainty, as the images resulting from moving the phantom are often equivalent to different initial conditions that fall within the training distribution. 

\subsubsection{Experiment 3.2: Moving the phantom after insertion}
\begin{figure}[h]
    \centering
    \includegraphics[width=\linewidth]{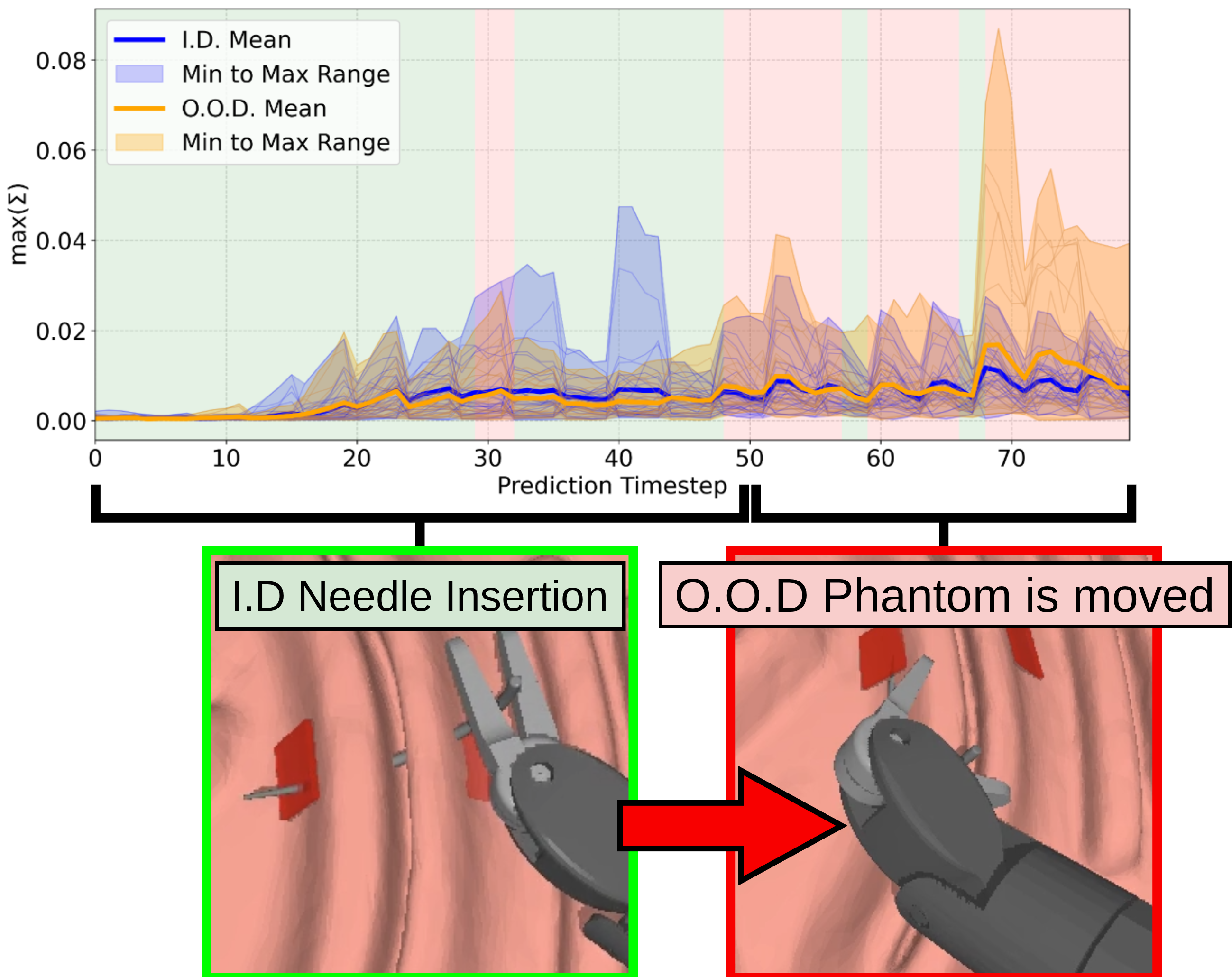}
    \caption{The top graph depicts the estimated uncertainty ($\max (\Sigma)$) for each prediction at each timestep by the model ensemble. The background indicates the prediction of the O.O.D. detector, i.e., green for I.D. and red for O.O.D. We execute the scenario 27 times. We use the maximum sample of the 27 runs during the O.O.D detection. The bottom images show that at the beginning of the scenario, the suture pad remains still; however, it is moved after needle insertion, causing an unseen state, which is detected around timestep 70.}
    \label{fig:UQ_Servo_last}
\end{figure}

The estimated uncertainty is illustrated in Fig.~\ref{fig:UQ_Servo_last}. The uncertainty increases significantly at the final step, where the phantom is moved after the needle has been inserted, creating a new state not encountered in the training data.

\subsection{Experiment 4: Control Barrier Functions}
\begin{figure}[h]
    \centering
    \includegraphics[width=\linewidth]{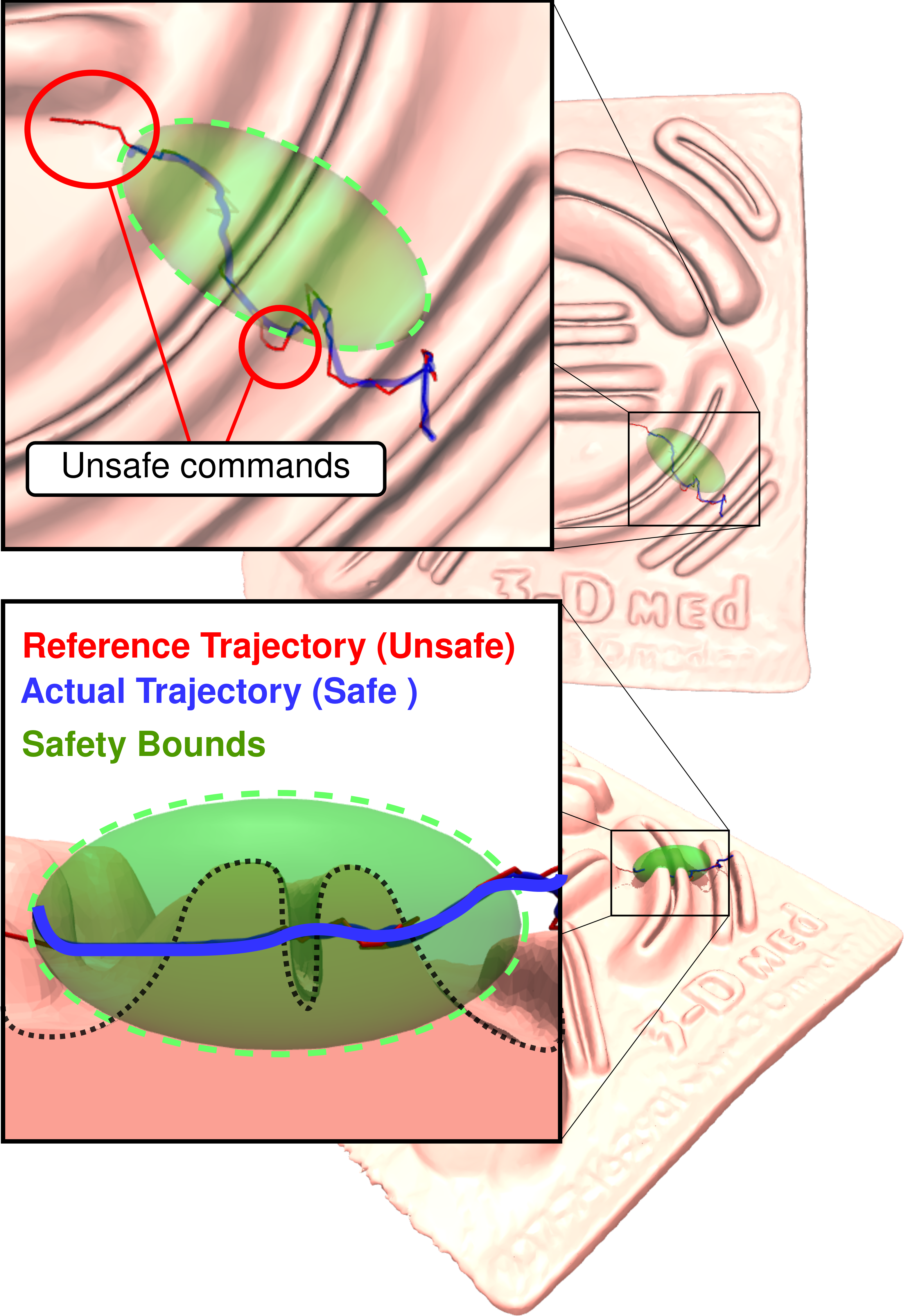}
    \caption{The manually-set safety bounds-ellipsoid around the suturing area is marked in green. The reference trajectory, i.e., the predicted action from the Diffusion Policy, which is potentially unsafe, is depicted in red. The actual trajectory -- i.e., the resulting safe trajectory from the CBF -- that is executed by the robot is shown in blue.}
    \label{fig:CBF_in_sim}
\end{figure}

\begin{figure*}
    \centering
    \includegraphics[width=\linewidth]{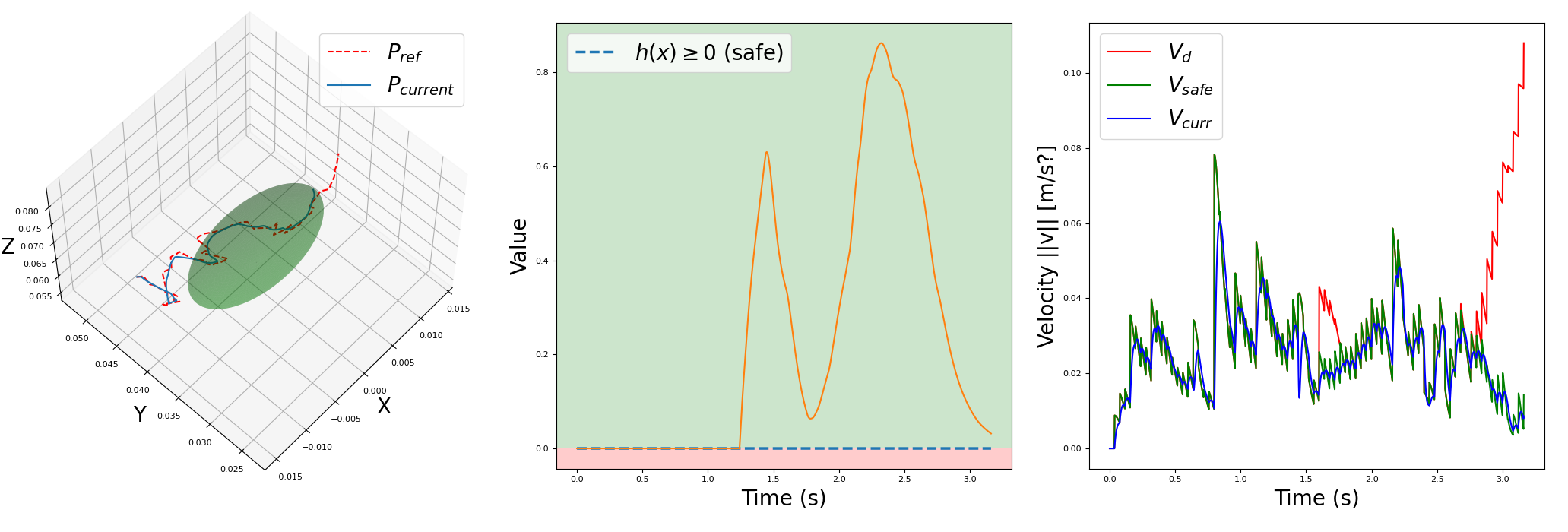}
    \caption{Left) 3D plot showing the reference trajectory (red), the actual trajectory after CBF (blue), and the ellipsoid safety bound (green) Middle) Value of the barrier function $h(x)$. A value $h(x)\geq0$ means that the system remains inside the safe region.  Right) Shows the desired (unsafe), safe (filtered by the CBF controller), and current (followed by the robot) velocities of the execution.}
    \label{fig:CBFs}
\end{figure*}

Due to technical limitations in the simulator (see \cref{sec:discussion}), it was not possible to implement our CBFs to run live on the system. For those reasons, we illustrate the CBFs offline. We run an execution in the simulator and record the needle tip position, which we use as our reference trajectory. A safety set has been defined as an ellipsoid at the center between the suturing entry and exit points. The CBF is enabled once the needle tip has entered the safety set, where the CBF limits actions to remain inside the safety set, as seen in Fig.~\ref{fig:CBF_in_sim}. Fig.~\ref{fig:CBFs} shows that the CBF approach is able to limit the controlled velocity $V_\text{curr}$ to ensure that safety is not violated, i.e., that $h(x) \geq 0$ for all time steps, whereas naively following the desired velocity $V_d$ would lead to unsafe positions.

\section{Discussion and Limitations}
\label{sec:discussion}
Our results demonstrate that through the adopted Diffusion Policy ensemble model, our method is able to detect a significant increase in the prediction uncertainty for all test cases where O.O.D. scenarios were induced.
For the test case with needle drop, experiment 1, it's possible to detect that the needle is dropped; however, not immediately. This could be due to the low visual disparity between I.D. and O.O.D. cases since the needle is occluded mainly by either phantom or PSM. 
As shown in our results in experiment 2, it is evident that the model uncertainty depends heavily on the visual features present in the image, as the uncertainty measures increase when changing the camera viewpoint changes. In many practical scenarios, the model should be viewpoint-independent, as moving the camera during procedures is common and essential. However, in our case, we simulate a poor initial viewing angle (O.O.D.) that differs significantly from the training dataset in contrast to an ideal viewing angle (I.D.). The ability of the model to detect a sub-optimal viewpoint based on the features in-view in the image shows promise for automating tasks such as Critical View of Safety in robotic laparoscopic cholecystectomy, where it is crucial to maintain visibility of certain anatomical landmarks \cite{manatakis2023critical}.

With regards to the policy's ability to generalize, it is interesting to note that, despite moving the phantom around during the suturing scenarios, as seen in experiment 3.1 -- which was not per se included in the training data -- the model was able to generalize to overcome this type of perturbation without it being categorized as O.O.D., as the training data included a variety of phantom positions. However, due to the dynamics of moving the phantom in the simulation, in cases where the movement was more abrupt or farther from the training set, an O.O.D. detection was triggered.
The same applies to moving the phantom after needle insertion has begun, experiment 3.2, as the dynamics of moving the phantom while the needle is inserted cause the PSM to reach unseen states.
Despite exhibiting increased levels of uncertainty during the test cases, the model shows corrective behavior, attempting to perform the needle insertion despite the camera or phantom being moved. This raises interesting perspectives on the tradeoff between O.O.D. detection for safety and the ability of the policy to generalize.

\subsection{Challenges of Sampling-based Uncertainty Quantification}
\begin{figure}[!h]
    \centering
    \includegraphics[width=0.8\linewidth]{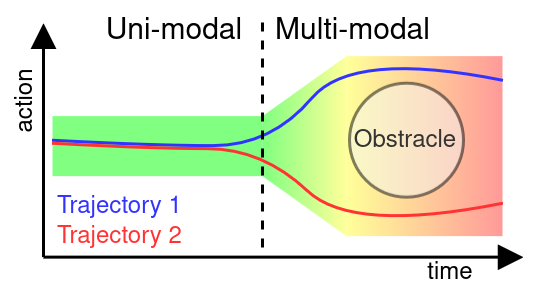}
    \caption{Illustrates a common occurrence of uni- and multi-modality in demonstration trajectories. Where two different demonstrations avoid obstacles by taking two different actions}
    \label{fig:uni-multi-dist}
\end{figure}

\begin{figure}[!h]
    \centering
    \includegraphics[width=0.8\linewidth]{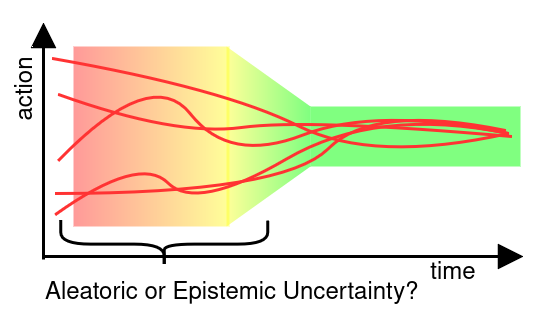}
    \caption{Illustrates a common occurrence where multiple demonstrations converge to a singular point, e.g., reaching an object from different starting conditions. This case highlights the difficulty of distinguishing between Aleatoric and Epistemic uncertainty at the beginning of the demonstration trajectory.}
    \label{fig:aleatoric-epistemic}
\end{figure}

An inherent challenge in uncertainty quantification and O.O.D. detection is the need to calibrate the distribution of the I.D. data, i.e., the null hypothesis in O.O.D. detection. This gives rise to two major challenges: 1) Handling Multi-modal distributions. 2) Distinguishing between epistemic and aleatory uncertainty. A schematic representation of these cases is shown in Fig.~\ref{fig:uni-multi-dist} and Fig.~\ref{fig:aleatoric-epistemic}.\\

\noindent
In Fig.~\ref{fig:uni-multi-dist}, the target distribution is initially uni-modal but changes to multi-modal throughout the policy rollout. This is a common case seen in demonstration data, as there can be multiple ways to reach the same position, e.g., one can go around an object by moving left or right, where both directions are considered equally valid. Conflating uni-modal and multi-modal distributions can lead to poorly calibrated uncertainty estimates. Further, as the model ensemble approach to uncertainty quantification is inherently sampling-based, correctly characterizing the distribution may require many samples, which can be computationally infeasible (as this requires running many models in parallel). 

In Fig.~\ref{fig:aleatoric-epistemic}, we see action trajectories that initially exhibit large variance, but later converge around a single point. This is a common occurrence in robot tasks, as there can be many valid trajectories to approach a single point, but the final point itself may be constrained, as is the case in suturing. In this case, whereas in the final segment of the motion near the target point the variance can likely be explained as aleatoric uncertainty (i.e., measurement noise), for the first part, it is unclear whether the high variance is due to aleatoric uncertainty -- i.e., many ways of completing the task -- or epistemic uncertainty -- i.e., the model ensemble is unsure of what to do. This poses challenges when calibrating the distribution of the I.D. uncertainty.

Both cases above are well-known problems in the uncertainty quantification machine learning literature \cite{dong2024multiood}\cite{osband2016risk}, and fall beyond the scope of the current paper. 

\subsection{Limitations of our CBF Integration}
A implemented in our current framework, the model-free CBF approach suffers from a number of limitations in practical use. Although the model-free approach we adopt circumvents the sensitivity of model-based methods to model uncertainties, model-free CBFs are still sensitive to external disturbances. These are included in our system through $\mathbf{d}$ in \cref{pre:robot_dynamics_dis}. However, the magnitude of $\mathbf{d}$ should in practice be chosen based on the maximum expected disturbance. This could, e.g., take into account kinematic error. It is commonly known that the forward kinematics of the PSM is inaccurate, since joint encoders are only present on the actuator side and the cable-driven mechanism will introduce slack. Further disturbance terms could be introduced to model the flexibility of the laparoscopic tools when in-contact with tissue.

A requirement to ensure safety-guarantees are met under the model-free CBF framework is that the velocity controller tracks the reference velocity fast enough \cite{molnar2021model}. In our evaluation, we tested the model-free CBF offline. This was due to limitations of the Asynchronous Multi-Body Framework (AMBF) used for simulation, where due to the design of the control interface used to command the robot and to the slow response of the actuators, it was not possible to implement a closed-loop velocity controller running faster than 1 Hz, whereas the DaVinci Research Development Kit can be commanded and reach targets at a minimum of 100 Hz \cite{yang2024effectiveness}. Thus, this is not expected to be a problem in practice. However, as explained above, the tendon-driven mechanisms in the PSM could introduce kinematic errors, leading to poor velocity tracking. This could be compensated by modeling the actuator hysteresis and should be investigated in future work.

In a clinical scenario, a practical way of defining the safe region enforced by the CBF must be devised. From a procedural perspective, this could be done at the pre-operative stage and applied using real-time computer vision registration, or could be manually defined by the surgeon through a Graphical User Interface. Other desired shapes of the CBF are possible and could be defined analytically to fit more tissue and wound types and shapes. If an arbitrary shape of the CBF without an analytical expression is desired, learning-based methods for CBF synthesis exist \cite{ChengRLCBF}.

\subsection{Incorporating the Surgeon}
Using a diffusion policy ensemble to quantify prediction uncertainty, we automatically detect when the model is operating I.D and O.O.D.. By detecting this, we can automatically relay system control back to the surgeon when the model encounters unseen situations where the risk of it making decision errors is higher. The surgeon then manually controls the system and performs corrective actions. This idea is related to the switch between manual and automatic control in Levels 2-3 of surgical robot autonomy \cite{attanasio2021autonomy}. Whereas the majority of prior approaches implement this switch purely based on the state of the task, we propose that this switch should also take into account the system's uncertainty in its autonomous decision-making.
An interesting direction would be to use the surgeon's performance of these corrective actions to further train the system. Additionally, utilizing the uncertainty measures, it becomes possible to localize specific scenarios where the uncertainty measures are large, meaning further expert demonstrations are needed. Similar human-in-the-loop and active learning strategies have been investigated within medical image analysis \cite{budd2021survey}.

\section{Conclusion}
\label{sec:conclusion}
This study proposed a novel framework for safe, uncertainty-aware imitation learning of RMIS suturing policies. We used an ensemble of Diffusion Policies to learn a needle insertion task by training on expert demonstrations, allowing us to quantify prediction uncertainty. We used the estimated uncertainty to create an out-of-distribution detector that can stop the system and relay control to the surgeon in unseen scenarios. Furthermore, model-free Control Barrier Functions were used as an additional layer of safety to place formal guarantees on the controller's safety. We evaluated our system by testing several cases, such as dropping the needle, moving the camera to a suboptimal and previously unseen view, and moving the phantom in an unexpected manner. Using the estimated uncertainty, it was possible to detect these out-of-distribution scenarios consistently during multiple executions. Additionally, the learned policy was robust to these perturbations, showing corrective behaviors and generalization. 
We also showed that in cases of unsafe predictions by our learned policy, the Control Barrier Function limits the action such that the needle tip remains within our specified safety set, adding a redundant layer of safety to our framework.

\bibliographystyle{ieeetr}
\bibliography{references}

\end{document}